\documentclass[sigconf]{acmart}
\AtBeginDocument{%
  \providecommand\BibTeX{{%
    \normalfont B\kern-0.5em{\scshape i\kern-0.25em b}\kern-0.8em\TeX}}}

\usepackage{multirow}
\usepackage{multicol}
\setcopyright{acmcopyright}

\copyrightyear{2022}
\acmYear{2022}
\setcopyright{acmcopyright}\acmConference[WWW '22 Companion]{Companion Proceedings of the Web Conference 2022}{April 25--29, 2022}{Virtual Event, Lyon, France}
\acmBooktitle{Companion Proceedings of the Web Conference 2022 (WWW '22 Companion), April 25--29, 2022, Virtual Event, Lyon, France}
\acmPrice{15.00}
\acmDOI{10.1145/3487553.3524238}
\acmISBN{978-1-4503-9130-6/22/04}

\DeclareUnicodeCharacter{FB01}{fi}

\usepackage{xcolor}
\usepackage{calc}
\let\mc\multicolumn
\newcommand{\ours}{\textbf{GenKGC}}
\begin{document}

\title{From Discrimination to Generation:  \protect\\ Knowledge Graph Completion with Generative Transformer}


\author{Xin Xie}
\affiliation{%
  \institution{Zhejiang University \& AZFT Joint Lab for Knowledge Engine}
  \city{Hangzhou}
  \country{China}
}
\email{xx2020@zju.edu.cn}

\author{Ningyu Zhang}
\authornotemark[1]
\affiliation{%
  \institution{Zhejiang University \& AZFT Joint Lab for Knowledge Engine}
  \city{Hangzhou}
  \country{China}
}
\email{zhangningyu@zju.edu.cn}

\author{Zhoubo Li, Shumin Deng}
\affiliation{%
  \institution{Zhejiang University \& AZFT Joint Lab for Knowledge Engine}
  \city{Hangzhou}
  \country{China}
}
\email{{zhoubo.li,231sm}@zju.edu.cn}

\author{Hui Chen}
\affiliation{%
  \institution{Alibaba Group}
  \city{Hangzhou}
  \country{China}
}
\email{weidu.ch@alibaba-inc.com}

\author{Feiyu Xiong, Mosha Chen}
\affiliation{%
  \institution{Alibaba Group}
  \city{Hangzhou}
  \country{China}
}
\email{{feiyu.xfy,chenmosha.cms}@alibaba-inc.com}

\author{Huajun Chen}
\authornote{Corresponding author.}
\affiliation{%
  \institution{Zhejiang University \& AZFT Joint Lab for Knowledge Engine}
  \city{Hangzhou}
  \country{China}
}
\email{huajunsir@zju.edu.cn}

\renewcommand{\shortauthors}{Xin Xie, et al.}

\begin{abstract}

Knowledge graph completion aims to address the problem of extending a KG with missing triples. In this paper, we provide an approach \textbf{GenKGC}, which converts knowledge graph completion to sequence-to-sequence generation task with the pre-trained language model. We further introduce relation-guided demonstration and entity-aware hierarchical decoding for better representation learning and fast inference. Experimental results on three datasets show that our approach can obtain better or comparable performance than baselines and achieve faster inference speed compared with previous methods with pre-trained language models. We also release a new large-scale Chinese knowledge graph dataset OpenBG500 for research purpose\footnote{Code and datasets are available in \url{https://github.com/zjunlp/PromptKG/tree/main/research/GenKGC}}.
\end{abstract}





\maketitle

\section{Introduction}

Knowledge Graphs (KGs) treat the knowledge in the real world as fact triples in the form of \texttt{<subject, predicate, object>}, abridged as $(s, p, o)$, where $s$ and $o$ denote entities and $p$ are the relations between entities, which can benefit a wide range of knowledge-intensive tasks.
Knowledge graph completion (KGC) aims to complete the knowledge graph by predicting the missing triples.
In this paper, we mainly target \textit{link prediction} task for KGC based on the powerful pre-trained language models.

Most previous KG completion methods, such as TransE \cite{Bordes:TransE}, ComplEx \cite{complex}, and RotatE \cite{RotatE}, are knowledge embedding techniques that embed the entities and relations into a vector space and then obtain the predicted triples by leveraging pre-defined scoring functions to those vectors.
Recently, some textual encoding methods (e.g., KG-BERT \cite{kgbert}) have been proposed, which utilize the pre-trained language model to encode triples and output the score for each candidate.
Obviously, most previous approaches leverage the discrimination paradigm with a pre-defined scoring function to knowledge embeddings. 
However, such a discrimination strategy has to costly scoring of all possible triples in inference and suffer from the instability of negative sampling. 
Moreover, those dense knowledge embedding approaches (e.g., TransE) ignore the fine-grained interactions between entities and relations and have to allocate a large memory footprint for the large-scale real-world knowledge graph. 
Therefore, it is intuitive to find a new technical solution for knowledge graph completion.

\begin{figure*}[!t] %H为当前位置，!htb为忽略美学标准，htbp为浮动图形
\centering %图片居中
\includegraphics[width=0.98\textwidth]{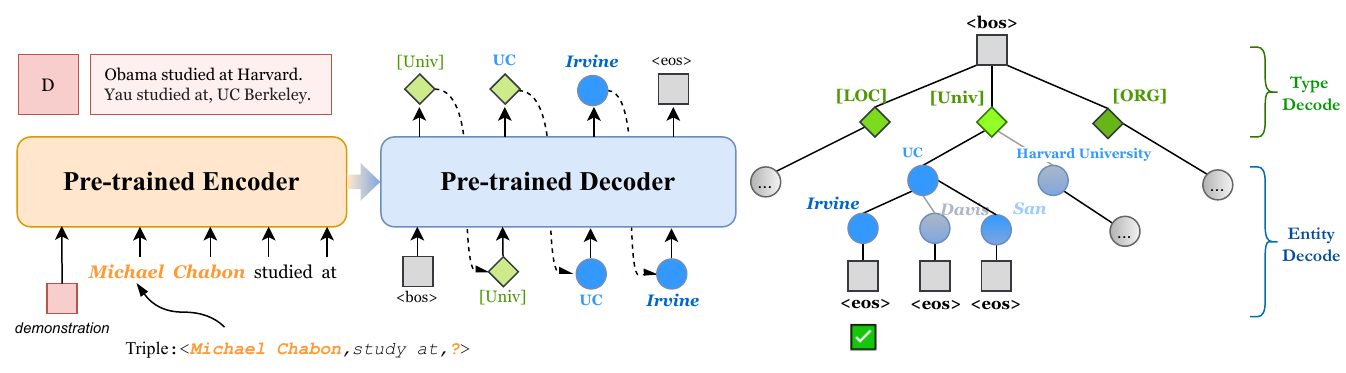} %插入图片，[]中设置图片大小，{}中是图片文件名
\caption{
Architecture of \ours. We augment the input text of entity and relation with  demonstrations, and introduce entity-aware hierarchical decoding for fast inference.}
\label{fig:model}
\end{figure*}

In this paper, we take the first step to model the knowledge graph completion with sequence to sequence generation and propose a novel approach~\ours.
To be specific, we represent entities and relations as input sequences and utilize the pre-trained language model to generate target entities.
Following GPT-3's naive "in-context learning" paradigm, in which the model can learn correct output answers by concatenating the selected samples relevant to the input, we propose relation-guided demonstration by adding triples of the same relation.
Moreover, during generation, we propose entity-aware hierarchical decoding to reduce the time complexity of generation.
Experimental results on two datasets WN18RR, FB15k-237 and a newly released large-scale Chinese KG dataset OpenBG500 demonstrate the effectiveness of the proposed approach.
The contributions of our work are as follows:

\begin{itemize}
    \item We convert link prediction to sequence to sequence generation and propose~\ours, which can reduce the inference time while maintaining the performance.
    \item We propose relation-guided demonstration and entity-aware hierarchical decoding, which can better represent entities and relations and reduce the time complexity of generation.
    \item We report results on two datasets and release a new large-scale KG dataset, OpenBG500, for research purposes.
\end{itemize}

\section{Method}

\subsection{Link Prediction as Seq2Seq Generation}

Knowledge graph is defined with entity categories and entity descriptions as a tuple $\mathcal{G}=(\mathcal{E}, \mathcal{R}, \mathcal{T}, \mathcal{C}, \mathcal{D})$, where $\mathcal{E}$ represents a set of entities, $\mathcal{R}$ represents relation types, $\mathcal{T}$ refers to a set of triples , $ \mathcal{C}$ refers to the entity categories and $ \mathcal{D}$ refers to the entity descriptions. 
For each triple $t \in \mathcal{T}$, there is the form $(e_i, r_j, e_k)$ where $e_i, e_k \in \mathcal{E}$ is the head and tail entity respectively. 
For each entity $e_i \in \mathcal{E}$, there exists category $c_i$ to define the $e_i$ and a text $d_i$ to describe $e_i$.
To complete missing triples in KGs, link prediction aims at predicting the tail entity given the head entity and the query relation, denoted by $(e_i, r, ?)$.

In this paper, we utilize the standard encoder-decoder architecture for sequence-to-sequence generation. 
Note that we regard each entity and relation as the sequence of tokens. 
Concretely, we follow \cite{kgbert} to use plain text to represent the entities and relations instead of unique embedding to bridge the gap between the triples in the knowledge graph and pre-trained language models.
To be specific, given a triple with tail entity missing $(e_i, r_j, ?)$, we obtain the description $d_{e_i}$ $d{r_j}$ of $e_i$ and $r_j$ respectively, and concatenate them together.
For example, as shown in Figure \ref{fig:model}, \textit{"Michael Chabon studied at"} (head entity, relation) is the major part of the input sequence, and \textit{"UC, Irvine"} (target entity) is part of the output sequence. 
Thus, we have the input sequence of $<e_i, r>$ pair and generate the output sequence of $e_k$. 
We leverage BART for model training and inference. 

\let\mc\multicolumn

\begin{table}[!htbp]

\centering
\caption{Inference and training efficiency comparison. $|d|$ is the length of the entity description. $|\mathcal{E}|$ is the numbers of all unique entities in the KG. 
$k$ is the negative samples for KG-BERT and the beam size for our \ours~.
The time under RTX 3090 is used to estimate the speed of training and inference given a single (entity,relation) pair on OpenBG500.
}
\setlength{\tabcolsep}{1mm}
\resizebox{.47\textwidth}{!}{% <------ Don't forget this %
\begin{tabular}{lccc}
\toprule
For One Triple & Method & Complexity & Time under RTX 3090\\
\midrule 
\multirow{3}{*}{Training}       & TransE    & $O(k+1)$    &    0.08ms    \\
                            & KG-BERT    & $O(|d|^{2} \times (k+1))$      &       72ms       \\
                                & \ours      & $O(|d|^{2})$             &       2.35ms         \\
\midrule 
\multirow{3}{*}{Inference}     & TransE    & $O(|\mathcal{E}|)$ &    0.02s   \\
                                & KG-BERT    & $O(|d|^{2} \times |\mathcal{E}|  )$ &    10100s  \\
                                & \ours      & $O(|d|^{2} \times |d|^{k})$  &      0.96s       \\
\bottomrule
\end{tabular}
}%
\label{tb:time}
\end{table}

\subsection{Relation-guided Demonstration}

Inspired by prompt-tuning \cite{DBLP:journals/corr/abs-2108-13161}, we propose relation-guided demonstration for the encoder. 
Note that there exist long-tailed distributions in the KGs, for example, there  exist only 37 instances of the relation \texttt{film/type\_of\_appearance} in the FB15k-237 dataset.
Previous study \cite{DBLP:journals/corr/abs-2108-13161} illustrates that concatenating  randomly sampled instances as demonstrations can yield better few-shot performance.
Thus, we construct relation-guided demonstration examples $ \{r_{\text{in}}, t_{\text{train}}\} $.
To be specific, we sample those demonstrations with the guidance of relation $r_j$, which consists of several triples with the same relation of input from training set.
Formally, we have the final input sequence as:
$$
x = \texttt{<bos>} \ \text{demonstration}(r_j) \ \texttt{<sep>} \ d_{e_i} \ d_{r_j} \  \texttt{<sep>}
$$

\begin{table*}[!ht]

\centering
\caption{Experiment results on WN18RR, FB15k-237 and OpenBG500.  $\diamond$Resulting numbers are reported by \protect\cite{nathani2019learning}, we reproduce the model result on OpenBG500 and take other results from the original papers.
% The bold numbers denote the best results in each genre while the underlined ones are the second-best performance.
}
\resizebox{0.8\textwidth}{!}{% <------ Don't forget this %
\begin{tabular}{lcccccccccc}
\toprule
              &         \mc{3}{c}{\bf WN18RR}                           &        \mc{3}{c}{\bf FB15k-237}        & \mc{3}{c}{\bf OpenBG500}  \\
                        \cmidrule(lr){2-4}                         \cmidrule(lr){5-7}              \cmidrule(lr){8-10}
Method        &    Hits@1   &     Hits@3  &     Hits@10     &    Hits@1 &   Hits@3  &  Hits@10    & Hits@1   &     Hits@3  &     Hits@10   \\ 
\midrule      
\multicolumn{10}{c}{\textit{Graph embedding approach}}                                                                              \\
\midrule
TransE \cite{Bordes:TransE}	$\diamond$    &   0.043	&   0.441	&   0.532	&   0.198	&   0.376	&   0.441  & 0.207 & 0.340 & 0.513 \\ 
DistMult \cite{distmult} $\diamond$	&   0.412	&   0.470	&   0.504	  &   0.199   &   0.301	&   0.446   & 0.049 & 0.088 & 0.216\\
ComplEx	\cite{complex} $\diamond$    &   0.409	&   0.469	&   0.530	&   0.194	&   0.297	&   0.450   & 0.053 & 0.120 & 0.266\\
RotatE \cite{RotatE}	    &   0.428	&   0.492	&   0.571		&   0.241	&   0.375	&   0.533 & - & - & -\\
TuckER \cite{tucker}	    &   0.443	&   0.482	&   0.526	&   0.226	&   0.394	&   0.544   & - & - & - \\
ATTH \cite{ATTH}	    &0.443	&  0.499	   &   0.486	&   0.252	&   0.384	&   0.549 & - & - & - \\
\midrule
\multicolumn{10}{c}{\textit{Textual encoding approach}}                                                                             \\
\midrule
KG-BERT	\cite{kgbert}    &   0.041	&   0.302	&   0.524		&   -	    &   -	    &   0.420   & 0.023 & 0.049 & 0.241\\
StAR \cite{STAR}	    &   0.243	&   0.491	& 0.709		&   0.205	&   0.322	&   0.482  & - & - &  -\\
\midrule
\ours	    & 0.287	&  0.403	&   0.535	& 0.192	& 0.355	& 0.439 & 0.203& 0.280& 0.351\\
\bottomrule
\end{tabular}
}%

\label{tb:transductive}%
\end{table*}

\subsection{Entity-aware Hierarchical Decoding}

In the vanilla decoding setting, we have to enumerate all entities in the $\mathcal{E}$ and then sort them by the score function.
However, this process can be time-consuming, as shown in  Table \ref{tb:time} when $\mathcal{E}$ is very large, it is costly scoring of all possible triples.
In our approach, we follow  \cite{DBLP:conf/iclr/CaoI0P21} to use the Beam Search to obtain top-$k$ entities in the $\mathcal{E}$ ($k$ is the hyperparameter of beam size).
Intuitively, there is no need for negative sampling as we directly optimize by predicting the correct entity in decoding.
To be more specific, given with a triple with tail or head entity missing $(e_i,r_j,?)$, \ours~  rank each $e \in \mathcal{E}$ by computing a score with an autoregressive formulation:

\begin{equation}
p_{\theta}(y \mid x)=\prod_{i=1}^{|c|} p_{\theta}\left(z_{i} \mid z_{<i}, x\right)  \prod_{i=|c|+1}^{N} p_{\theta}\left(y_{i} \mid y_{<i}, x\right) ,
\end{equation}
where $z$ is the set of $|C|$ tokens in the category $c$, $y$ is the set of $N$ tokens in the textual representation of $e$. %and $\theta$ is the parameters of the model.

Since KG contains rich semantic information such as entity types, it is intuitive to constrain the decoding for fast inference.
We sample the type category with the lowest frequency of occurrence in the training set to constrain the entity decoding since it is challenging to discriminate those low-frequent entities.
Then, we add special tokens as types in the  vocabulary of pre-trained language model to constrain the decoding. 
To ensure that the generated entities are inside the entity candidate set, we construct a prefix tree (trie tree) to decode our entity name as shown in Figure \ref{fig:model}.
Similar to the ordinary sequence-to-sequence model, we optimize \ours~using a standard seq2seq objective function:

\begin{equation}
\mathcal{L} = - \log p_{\theta}(y \mid x)
\end{equation}

\section{Experiment}

\paragraph{Datasets}
We evaluate our method on FB15k-237 \cite{fb15k}, WN18RR \cite{wn18rr}, which are widely used in the link prediction literature, and a new real-world large-scale Chinese KG dataset OpenBG500\footnote{OpenBG500 is a subset of open business KG from \url{https://kg.alibaba.com/}.}.
In FB15k-237, descriptions of entities are obtained from the introduction section of the Wikipedia page of each entity. 
In WN18RR, each entity corresponds to a word sense, and description is the word definition. 
In OpenBG500, the descriptions of entities and relations come from the e-commerce description page.
More details about datasets are listed in Table \ref{tb:dataset}.

\begin{table}[!htbp]\small
\caption{\small Summary statistics of benchmark datasets.}
\renewcommand\tabcolsep{4.0pt}
	\centering
	\begin{tabular}{lccccc}
		\toprule
		\textbf{Dataset}   & \textbf{\# Ent}  & \textbf{\# Rel} & \textbf{\# Train} & \textbf{\# Dev} & \textbf{\# Test} \\
		\midrule
		WN18RR    & 40,943  & 11    & 86,835   & 3,034  & 3,134   \\
		FB15k-237 & 14,541  & 237   & 272,115  & 17,535 & 20,466  \\ 
		OpenBG500  & 269,658  & 500   & 1,242,550  & 5,000  & 5,000 \\
		\bottomrule
	\end{tabular}
	\label{tb:dataset}
\end{table}

\paragraph{Setting}
We adopt BART-base as our backbone for comparison with other BERT-based KGC methods like KG-BERT.
Following StAR \cite{STAR}, we choose the two kinds of KGE methods as our baseline models, which can be classified as graph embedding approach and textual encoding approach.
Grid search is used over the three datasets and the results are reported on the test set with hyperparameters of the best performance determined by the dev set.

\paragraph{Metrics}
We evaluate the test set of triples $\mathcal{T'}$ disjoint from the set of training triples $\mathcal{T}$.
For inference on a test triple $(e_i, r_j, e_k)$, we make sure predict the entity in the candidate set $\mathcal{E}$.
We use the metrics of hits@1, hits@3 and hits@10.
% The entities in the test triple are assumed to be seen in the training set.

\paragraph{Comparison with Discrimination Method}

\definecolor{color1}{rgb}{0.22,0.45,0.70}  % light blue
\definecolor{color2}{rgb}{0.45,0.45,0.45}  % dark grey

\newcommand{\progressbar}[2][2cm]{%
    \textcolor{color1}{\rule{#1 * \real{#2} / 100}{1.5ex}}%
    \textcolor{color2!15}{\rule{#1 - #1 * \real{#2} / 100}{1.5ex}}}

\begin{table}[!htb]

\centering
\caption{We list a query and first five entities with their probability predicted by \ours~w/o entity-aware decoding, and its reranking with \ours.}
\resizebox{0.42\textwidth}{!}{
\begin{tabular}{cll}
\toprule
\mc{3}{l}{\textbf{Query:}\texttt{(?,student,Michael Chabon)}}      \\
\midrule
Rank & \ours~w/o hierarchical decoding        &Probability                   \\
\midrule
% University of California, San Francisco
% 1135
% /m/02_xgp2	/education/educational_degree/people_with_this_degree./education/education/institution	/m/07vfz
1 & University of California            & \hfill \progressbar{42}          \\
2 & \textbf{University of California, Irvine} & \hfill \progressbar{20} \\
3 & University of California, San Francisco & \hfill \progressbar{14}  \\
4 & University of California, Davis & \hfill \progressbar{12} \\
5 & University of California, Santa Cruz & \hfill \progressbar{9}\\
\midrule
Rank &  \ours  &Probability\\
\midrule
1 &\textbf{University of California, Irvine}  & \hfill \progressbar{25}     \\
2 &University of California, San Francisco   & \hfill \progressbar{24}   \\
3 &University of California, Davis          & \hfill \progressbar{17}      \\
4 &University of California, Santa Cruz& \hfill \progressbar{14}  \\
5 &University of Calgary        & \hfill \progressbar{11}      \\
\bottomrule
\end{tabular}
}
\label{tb:case study}
\end{table}

From Table \ref{tb:transductive}, we notice that \ours~ achieves better performance than KG-BERT \cite{kgbert} across all datasets and maintains high speed during inference.
The translation-based method like TransE, which treats entities and relations as dense vectors in the same space, will face the memory explosion problem.
For the memory cost, TransE has to consume 260M parameters to store the entities and relations for OpenBG500 with more than 260k entities,  
while pre-trained models (BERT or BART) only utilize 110M parameters, which demonstrates the memory efficiency of our approach.
Note that this problem will be more severe when the entities become more numerous because the space complexity is $O(n)$.
For the inference speed, KG-BERT encodes the relational triples with the pre-trained language model and ranks all candidate triples with correct and wrong entities for inference.
When the candidate entities set is huge, it is time-consuming for inference; for example, as shown in Table \ref{tb:time}, KG-BERT takes about 100100s to the scoring of all possible triples given a single (entity, relation) pair. 
While our method only needs to generate the top-$k$ entities with entity-aware hierarchical decoding, which \textbf{reduces lots of computing resources}.

\subsection{Case Study}
For different decoding strategies we conduct case study to analyze the result.
For \ours~w/o hierarchical decoding, we utilize the normal beam search to decode the text name of the missing entity.
From Table \ref{tb:case study}, we observe that \ours~ obtain better entity generation results while in normal beam search, the model may stop early at correct but not precise enough answer.
We argue that this is caused by the bias of the pre-trained language model (e.g., common token bias) since high-frequent tokens will lead the pre-trained language model to be biased toward certain answers.
Our entity-aware hierarchical decoding can constrain the decoding process and mitigate the bias effect caused by pre-trained language models.

\section{Related Work}
\paragraph{Knowledge Graph Embedding Models}
There are lots of methods of KGC are based on knowledge graph embeddings (KGE), which generally leverage an embedding vector in the continuous embedding space to represent the entity and the relation in KG \cite{DBLP:conf/www/ZhangDSCZC20}.
One kind of KGE methods is \textit{translation-based}, such as TransE \cite{Bordes:TransE},  ConE \cite{zhang2021cone}, TotatE \cite{RotatE},  which consider relations as the mapping function between entities.% by using distance-based scoring functions.
% use distance-based scoring functions  considering relations as the translational operation between entities.
% The objective of \textit{translation-based} methods is that the translated head entity should be close to the tail entity in real space \cite{Bordes:TransE}, complex space \cite{RotatE}, and have shown state-of-the-art performance on handling multiple relation patterns.
The other kind of KGE method is \textit{semantic matching} models, where they get the semantic similarity by using the multi-linear or bilinear product.

\paragraph{Pre-trained Language Models for KGC}
Recently, since pre-trained language models, such as BERT \cite{devlin2018bert}, have shown significant improvement on several natural language processing tasks, several works use the transformer-based models to tackle the KGC problems \cite{DBLP:journals/corr/abs-2201-05575,DBLP:journals/corr/abs-2201-11332,DBLP:journals/corr/abs-2201-11147}. 
KG-BERT \cite{kgbert} first propose to use BERT for KGC by seeing a triple as a sequence and converts KGC into a sequence classification problem with the binary cross-entropy object.
\cite{DBLP:journals/corr/abs-2112-08340} proposes to use a transformer encoder-decoder model that takes plain text as input and output a structured triple of the information hide in it.
% \citet{DBLP:conf/coling/KimHKS20} propose to adopt the multi-task learning framework with two additional tasks, i.e., relation prediction and relevance ranking, to improve high-rank performance based on KG-BERT. 

\section{Conclusion}

In this paper, we propose \ours, which can reach comparable results while reducing inference and training cost in link prediction with pre-trained models.
Experimental results on three benchmark datasets demonstrate the effectiveness of our approach, especially in inference time. 
%The success of \ours~suggests that link prediction as sequence-to-sequence generation can be helpful because entities can be inferred directly rather than costly scoring of all possible triples. 
%Future work should explore how to obtain the suitable categories for each entity in decoding. 

\begin{acks}
This work is funded by NSFC91846204/NSFCU19B2027, National Key R\&D Program of China (Funding No.SQ2018YFC000004), Zhejiang Provincial Natural Science Foundation of China  (No. LGG22F030011), Ningbo Natural Science Foundation (2021J190), and Yongjiang Talent Introduction Programme (2021A-156-G). 
\end{acks}

\bibliographystyle{ACM-Reference-Format}
\bibliography{sample-base}

\appendix

\end{document}